\documentclass[11pt,twoside]{article}

\usepackage[a4paper,top=2.6cm,bottom=2.8cm,left=2.8cm,right=2.8cm,headsep=0.8cm]{geometry}
\usepackage[T1]{fontenc}
\usepackage[utf8]{inputenc}
\usepackage{amsmath}
\usepackage{libertinus}
\usepackage{libertinust1math}
\usepackage[scaled=0.95,varqu,var0]{inconsolata}
\usepackage[final]{microtype}
\usepackage[table]{xcolor}
\definecolor{accent}{HTML}{6B2D5C}
\definecolor{accentlink}{HTML}{1F4E8C}
\definecolor{rulegray}{gray}{0.55}

\usepackage{titlesec}
\titleformat{\section}{\sffamily\Large\bfseries}{\textcolor{accent}{\thesection}}{0.8em}{}
\titleformat{\subsection}{\sffamily\large\bfseries}{\textcolor{accent}{\thesubsection}}{0.8em}{}
\titleformat{\paragraph}[runin]{\sffamily\normalsize\bfseries}{}{0pt}{}
\titlespacing*{\section}{0pt}{3.2ex plus 1ex minus .2ex}{1.4ex plus .2ex}
\titlespacing*{\subsection}{0pt}{2.6ex plus 1ex minus .2ex}{1.1ex plus .2ex}
\titlespacing*{\paragraph}{0pt}{1.6ex plus .6ex minus .2ex}{0.8em}

\usepackage[font=small,labelfont={sf,bf,color=accent},labelsep=period,margin=0.5cm]{caption}
\usepackage{fancyhdr}
\fancypagestyle{plain}{\fancyhf{}\fancyfoot[C]{\small\sffamily\thepage}}

\newif\ifblind
\newcommand{\weavecite}{\citep{campagne2026weave}}
\usepackage{tabularx}
\usepackage{graphicx}
\usepackage{booktabs}
\usepackage{url}
\usepackage{xurl}
\usepackage{listings}
\usepackage{tikz}
\usetikzlibrary{arrows.meta,positioning,fit,backgrounds}
\usepackage{placeins}
\graphicspath{{figures/}}
\usepackage{natbib}
\bibpunct[, ]{(}{)}{,}{a}{}{,}

\newlength{\figcol}
\usepackage{hyperref}
\hypersetup{
  colorlinks=true, breaklinks,
  linkcolor=accentlink, citecolor=accentlink, urlcolor=accent,
  pdftitle={GraphToolbox: A Configurable Framework for Controlled Benchmarking of Graph Neural Network Forecasting},
  pdfauthor={Eloi Campagne, Yvenn Amara-Ouali, Yannig Goude, Argyris Kalogeratos},
}

\definecolor{codebg}{RGB}{238,241,244}
\newcommand{\code}[1]{{\setlength{\fboxsep}{1pt}\colorbox{codebg}{\ttfamily\vphantom{Ag}#1}}}
\definecolor{figInk}{HTML}{0B0B0B}
\definecolor{figBlue}{HTML}{2A78D6}
\definecolor{figOrange}{HTML}{EB6834}
\definecolor{figGreen}{HTML}{1BAF7A}
\definecolor{codeBg}{HTML}{F6F6F3}
\definecolor{codeHl}{HTML}{FCEBD9}

\definecolor{vscKeyword}{HTML}{AF00DB} 
\definecolor{vscType}{HTML}{267F99} 
\definecolor{vscFunc}{HTML}{795E26} 
\definecolor{vscString}{HTML}{A31515} 
\definecolor{vscComment}{HTML}{008000} 
\definecolor{vscVar}{HTML}{001080} 
\definecolor{vscNum}{HTML}{098658} 

\lstdefinestyle{gtb}{
  language=Python,
  basicstyle=\ttfamily\scriptsize,
  keywordstyle=\color{vscKeyword},
  commentstyle=\color{vscComment}\itshape,
  stringstyle=\color{vscString},
  emph={[1]GATConv,myGNN,Trainer,ConvAdapter,ConvAdapterTemporal,GraphDataset,DataClass,%
    GraphBuilder,RollingTrainer,Optimizer,Aggregation},
  emphstyle={[1]\color{vscType}\bfseries},
  emph={[2]train,evaluate},
  emphstyle={[2]\color{vscFunc}},
  emph={[3]torch_geometric,nn,conv,graphtoolbox,models,gnn,training,%
    trainer,model,ds_train,ds_val,ds_test,pred,target,edge_index,attn,%
    in_channels,num_layers,hidden_channels,out_channels,conv_class,%
    conv_kwargs,batch_size,model_kwargs,patience,heads},
  emphstyle={[3]\color{vscVar}},
  literate=%
    {0}{{\textcolor{vscNum}{0}}}{1}{1}{{\textcolor{vscNum}{1}}}{1}%
    {2}{{\textcolor{vscNum}{2}}}{1}{3}{{\textcolor{vscNum}{3}}}{1}%
    {4}{{\textcolor{vscNum}{4}}}{1}{5}{{\textcolor{vscNum}{5}}}{1}%
    {6}{{\textcolor{vscNum}{6}}}{1}{7}{{\textcolor{vscNum}{7}}}{1}%
    {8}{{\textcolor{vscNum}{8}}}{1}{9}{{\textcolor{vscNum}{9}}}{1},
  showstringspaces=false,
  columns=fullflexible,
  breaklines=true,
  frame=tb,
  framerule=0.4pt,
  rulecolor=\color{figInk!35},
  backgroundcolor=\color{codeBg},
  framesep=4pt,
  framexleftmargin=4pt,
  framexrightmargin=4pt,
  xleftmargin=4pt,
  xrightmargin=4pt,
  escapeinside={(*@}{@*)},
  aboveskip=4pt,
  belowskip=2pt
}

\newcommand{\aff}[1]{\textsuperscript{\,#1}}
\makeatletter
\renewcommand{\maketitle}{%
  \thispagestyle{plain}%
  \begingroup\centering
  {\sffamily\LARGE\bfseries\color{accent}GraphToolbox\par}\vspace{1ex}
  {\sffamily\Large\color{accent}A Configurable Framework for Controlled Benchmarking\\[0.2ex]
   of Graph Neural Network Forecasting\par}\vspace{3ex}
  {\large\mbox{Eloi Campagne\aff{1,2,*}}\quad
   \mbox{Yvenn Amara-Ouali\aff{2,3}}\quad
   \mbox{Yannig Goude\aff{2,3}}\quad
   \mbox{Argyris Kalogeratos\aff{1,*}}\par}\vspace{1.4ex}
  {\small\itshape
   \aff{1}Centre Borelli, École Normale Supérieure Paris-Saclay, Gif-sur-Yvette, France\\
   \aff{2}EDF Lab, Palaiseau, France\\
   \aff{3}Laboratoire de Mathématiques d'Orsay, Université Paris-Saclay, Orsay, France\\[0.6ex]
   \upshape\aff{*}Corresponding authors:\\
   \href{mailto:eloi.campagne@ens-paris-saclay.fr}{\texttt{eloi.campagne@ens-paris-saclay.fr}},\quad
   \href{mailto:argyris.kalogeratos@ens-paris-saclay.fr}{\texttt{argyris.kalogeratos@ens-paris-saclay.fr}}\par}\vspace{3.5ex}
  \endgroup
  \setcounter{footnote}{0}%
}
\makeatother
\renewenvironment{abstract}{%
  \par\noindent{\color{rulegray}\rule{\linewidth}{0.4pt}}\par\vspace{1ex}
  \begin{list}{}{\leftmargin=1.2cm\rightmargin=1.2cm}\item\relax\small
  {\sffamily\bfseries\color{accent}Abstract.}\enspace\ignorespaces}{%
  \end{list}}
\newcommand{\keywords}[1]{%
  \begin{list}{}{\leftmargin=1.2cm\rightmargin=1.2cm}\item\relax\small
  {\sffamily\bfseries\color{accent}Keywords.}\enspace #1\end{list}
  \vspace{0.6ex}\noindent{\color{rulegray}\rule{\linewidth}{0.4pt}}\par\vspace{2ex}}

\begin{document}

\maketitle

\begin{abstract}
Electricity forecasting often involves spatially related signals observed over regions, substations, and feeders, and Graph Neural Networks (GNNs) represent these relations through a graph. 
Building a complete GNN forecasting experiment is nonetheless laborious, because graph construction, model selection, training, aggregation, and interpretation often require separate tools. We present GraphToolbox, an open-source Python framework that unifies these stages in one configuration-driven pipeline built on PyTorch Geometric. It offers data-driven graph construction, a single interface for most graph convolution operators of PyTorch Geometric and for recurrent graph cells, online expert aggregation, forecasting interpretability, and significance testing on forecasts. 
We use GraphToolbox to run a controlled benchmark: we compare architectural and modeling choices for French electricity load and net-load forecasting at regional and national levels under a common experimental protocol. On load, most evaluated graph convolutions outperform additive, recurrent, and foundation-model baselines. Pooled XGBoost, trained on the same inputs, has a lower RMSE than the best graph models, although the difference is not significant. Online expert aggregation and daily model updates improve accuracy further. On net-load, graph models outperform additive, recurrent, and XGBoost baselines, with further gains from forecasting the physical components separately. Statistical tests do not identify a single best convolution for load forecasting.
\end{abstract}


\section{Introduction}
\label{sec:introduction}
The operation of modern power systems rests on accurate short-term forecasts of electricity demand, which inform market decisions, unit commitment, and grid balancing. Historically, forecasters addressed this task at the level of an aggregated national signal, where generalized additive models (GAMs) and, more recently, boosting and deep learning methods have proven effective \citep{fasiolo2021fast}. Spatial dependencies between regions remain relevant when forecasts are aggregated at the national level. Electricity demand is distributed across interconnected regions, substations, and feeders whose components co-vary in ways that a single aggregated series cannot express. The decentralization of production, the integration of intermittent renewables, and the deployment of smart metering make this structure increasingly observable \citep{devilmarest2023adaptive}. \looseness=-1

\begin{figure*}[!htbp]
\centering
\hypersetup{hidelinks}
\newcommand{\figsub}{\tiny\color{figInk!70}}
\resizebox{\textwidth}{!}{\begin{tikzpicture}[
  font=\scriptsize, node distance=5mm,
  stage/.style={draw=figInk!55, line width=0.5pt, rounded corners=2pt, fill=figBlue!7, align=center,
    text width=21mm, minimum height=10.5mm, inner sep=1.5pt},
  cfg/.style={draw=figOrange!80, dashed, line width=0.5pt, rounded corners=2pt, fill=figOrange!5, align=center,
    text width=21mm, minimum height=6mm, inner sep=1.5pt, font=\tiny},
  outbox/.style={draw=figGreen!70!black, line width=0.5pt, rounded corners=2pt, fill=figGreen!8, align=center,
    text width=17mm, minimum height=5mm, inner sep=1pt},
  arr/.style={-{Stealth[length=1.6mm]}, line width=0.6pt, draw=figInk!80},
  badge/.style={circle, fill=figInk!75, text=white, font=\fontsize{4.5}{5}\selectfont\bfseries, inner sep=0pt,
    minimum size=3mm},
  carr/.style={-{Stealth[length=1.4mm]}, line width=0.5pt, draw=figOrange!80!black}
]
\node[stage] (data) {\texttt{DataClass}\\[-0.5pt]{\figsub CSV, splits}};
\node[stage, right=of data] (builder) {\texttt{GraphBuilder}\\[-0.5pt]{\figsub graph construction}};
\node[stage, right=of builder] (ds) {\texttt{GraphDataset}\\[-0.5pt]{\figsub scaling, lags}};
\node[stage, right=of ds] (model) {\texttt{myGNN}\\[-0.5pt]{\figsub \texttt{ConvAdapter}}\\[-2pt]{\figsub \texttt{ConvAdapterTemporal}}};
\node[stage, right=of model] (train) {\texttt{Trainer}\\[-0.5pt]{\figsub \texttt{RollingTrainer}}};
\foreach \a/\b in {data/builder, builder/ds, ds/model, model/train} \draw[arr] (\a) -- (\b);

\coordinate (bus) at ([xshift=4.5mm]train.east);
\node[outbox, anchor=west] (agg) at ([xshift=3.5mm, yshift=9.75mm]bus) {aggregation};
\node[outbox, anchor=west] (metrics) at ([xshift=3.5mm, yshift=3.25mm]bus) {metrics};
\node[outbox, anchor=west] (interp) at ([xshift=3.5mm, yshift=-3.25mm]bus) {interpretability};
\node[outbox, anchor=west] (signif) at ([xshift=3.5mm, yshift=-9.75mm]bus) {significance};
\draw[line width=0.6pt, draw=figInk!80] (train.east) -- (bus);
\draw[line width=0.6pt, draw=figInk!80] (bus |- agg.west) -- (bus |- signif.west);
\foreach \o in {agg, metrics, signif, interp} \draw[arr] (bus |- \o.west) -- (\o.west);

\node[cfg, below=4.5mm of data] (cfg0) {\texttt{data\_kwargs}};
\node[cfg, below=4.5mm of ds] (cfg1) {\texttt{dataset\_kwargs}};
\node[cfg, below=4.5mm of model] (cfg2) {\texttt{conv\_class}, \texttt{conv\_kwargs}};
\node[cfg, below=4.5mm of train] (cfg3) {\texttt{model\_kwargs},\\[-1pt] Optuna search space};
\foreach \n/\l/\s in {data/A/data, builder/A/data, ds/A/data, model/B/models, train/C/training, agg/C/training,
    metrics/C/training, interp/D/interp, signif/E/signif}
  \node[badge] at ([xshift=-0.4mm, yshift=-0.4mm]\n.north east) {\hyperref[sec:\s]{\l}};
\foreach \c/\t in {cfg0/data, cfg1/ds, cfg2/model, cfg3/train} \draw[carr] (\c) -- (\t);
\end{tikzpicture}}
\caption{Overview of the GraphToolbox pipeline. Solid arrows carry data from raw series to forecasts and diagnostics, while dashed boxes represent the configuration dictionaries that parameterize each stage. Swapping the graph, the convolution operator, or the training regime amounts to editing a configuration field. The badge of each module gives the subsection of Section~\ref{sec:design} that describes it.}
\label{fig:pipeline}
\end{figure*}

Graph Neural Networks build this structure into the model. By restricting information exchange to graph neighborhoods, they encode the assumption that spatial or statistical proximity governs load co-variation, an assumption that has driven their success in traffic forecasting and, increasingly, in power systems \citep{guo2019attention,campagne2024leveraging}. An empirical comparison nevertheless requires several methodological and software choices. A practitioner must first decide how to turn a collection of time series into a graph, then choose among a rapidly growing family of convolution operators, integrate those operators into a training loop with appropriate metrics, combine several models to gain robustness, and finally interpret the resulting predictions. A different library supports each of these steps, or none does, and the integration code is often specific to a single project.

GraphToolbox provides a common interface for these stages. The framework is an open-source Python package that organizes the entire forecasting workflow behind a small number of configuration dictionaries. Changing the graph, the convolution, or the aggregation strategy then requires modifying a configuration field while keeping the rest of the pipeline fixed. The same mechanism supports controlled comparisons, in which one choice varies (the convolution, the graph, the aggregation rule, the reconciliation, or the information given to the model) while the data, the split, and the training budget stay fixed.

\paragraph{Software contributions.} The framework's contributions are:
\begin{itemize}
\item It provides $6$ methods for constructing graphs from spatial coordinates or load measurements, and allows users to select a graph based on validation performance.
\item It provides a common forecasting interface for $53$ of the $65$ convolutional operators available in PyTorch Geometric (PyG), as well as recurrent graph cells from PyTorch Geometric Temporal (PyG-T). This makes it possible to compare these architectures without implementing a separate forecasting model for each one.
\item It combines the forecasts cached during an architectural sweep using online expert aggregation, and provides pairwise Diebold--Mariano tests, block-bootstrap intervals, and the Model Confidence Set for their statistical comparison.
\item It offers forecasting interpretation through accumulated local effects and an additive graph model for feature attribution. It also visualizes edge attributions, obtained from attention weights or occlusion-based influence, on a geographical graph.
\end{itemize}
\paragraph{Empirical findings.} We benchmark French electricity load and net-load forecasting across the $12$ regions, evaluating the sum of the regional forecasts at the national level. The main findings are as follows.
\begin{itemize}
\item The $46$ convolutions of the load sweep fall within $0.4$ points of mean absolute percentage error (MAPE) of each other, and the significance tests single out no best operator. Message passing accounts for part of their advantage over the additive, recurrent and foundation references, and recurrent graph cells do not improve on them. XGBoost pooled over the regions and given the same inputs has a lower RMSE than the best operator, a difference that is not significant.
\item Online aggregation outperforms the best individual convolution on both tasks and in every region for load forecasting, avoiding the need to choose an operator in advance. Updating one graph model daily also brings its accuracy in line with that of an additive model adapted through a Kalman filter.
\item On net-load, forecasting load, wind and solar separately improves every evaluated operator, and the best decomposed graph model is more accurate than the additive and recurrent references in every month and at every half-hour of the day. Given the same inputs, pooled XGBoost stays $433$~MW above the best direct graph model.
\end{itemize}

\section{Related tools}\label{sec:related}
PyG supplies convolution operators and the sparse primitives of message passing \citep{fey2019fast}, and PyG-T adds recurrent spatio-temporal cells and models built from them \citep{rozemberczki2021pytorch}. GraphToolbox builds on these libraries without replacing their implementations. Users specify a convolution or recurrent cell in the configuration, and the framework incorporates it into a common forecasting model that covers graph construction, training, and prediction. GraphToolbox fixes how these blocks assemble into a forecasting experiment, which the lower-level libraries leave open.

Closest in spirit is Torch Spatiotemporal (tsl), which provides data processing and model prototyping for neural forecasting over sensor networks \citep{cini2022tsl}. tsl and PyG-T remain general, supplying layers and data structures for arbitrary spatio-temporal signals, whereas GraphToolbox makes graph construction, architectural benchmarking, aggregation and interpretability explicit stages of one workflow. Toolkits built around global or foundation models forecast collections of series without a relational structure \citep{ansari2024chronos}. Table~\ref{tab:tools} compares GraphToolbox with PyG-T and tsl on these stages.

We score a capability as built-in, partial, or absent by whether a released module provides it directly. The adapter row requires instantiating and training an arbitrary PyG operator without operator-specific code, which the compared libraries offer only for a curated subset of layers. Data-driven graph construction requires deriving the adjacency from the series, so a library that accepts only a supplied or coordinate graph scores absent. Rolling-origin evaluation requires an expanding-window trainer exposed as a module. For interpretability, we consider accumulated local effects, the additive graph model, and the visualization of edge weights on a map. GraphToolbox can display attributions obtained from an external explainer but does not fit that explainer itself. The final group of rows identifies functionality available as dedicated modules in PyG-T or tsl but not in GraphToolbox, including ready-to-use datasets, curated spatio-temporal architectures, and imputation. These features are planned for a future release.

\begin{table}[!htbp]
\caption{Design emphasis of GraphToolbox relative to PyG-T and tsl, scored against the released modules of each library as of September 2026. The comparison is limited to the main design goals of each library.}
\label{tab:tools}
\centering
\footnotesize
\begin{tabular}{@{}lccc@{}}
\toprule
\textbf{Capability} & \textbf{GraphToolbox} & \textbf{PyG-T} & \textbf{tsl} \\
\midrule
\multicolumn{4}{@{}l}{\textit{Shared foundation}}\\
Built on PyG & \checkmark & \checkmark & \checkmark \\
\midrule
\multicolumn{4}{@{}l}{\textit{GraphToolbox emphasis}}\\
Adapter over arbitrary convs & \checkmark & $\sim$ & $\sim$ \\
Data-driven graph construction & \checkmark & -- & $\sim$ \\
Empirical graph selection & \checkmark & -- & -- \\
Online expert aggregation & \checkmark & -- & -- \\
Forecasting interpretability & \checkmark & -- & -- \\
Forecast significance testing & \checkmark & -- & -- \\
Rolling-origin evaluation & \checkmark & $\sim$ & \checkmark \\
Integrated hyperparameter search & \checkmark & -- & $\sim$ \\
\midrule
\multicolumn{4}{@{}l}{\textit{Complementary strengths}}\\
Prebuilt benchmark datasets & $\sim^{\dagger}$ & \checkmark & \checkmark \\
Curated ST model zoo & $\sim^{\dagger}$ & \checkmark & \checkmark \\
Missing-data imputation & --$^{\dagger}$ & -- & \checkmark \\
\bottomrule
\end{tabular}

\vspace{0.5em}
\footnotesize
\checkmark: built-in capability,\quad
$\sim$: partial capability,\quad
--: no relevant capability,\\
$\dagger$: capability planned for the next GraphToolbox release.
\end{table}

\section{Framework design}\label{sec:design}
GraphToolbox consists of modules that mirror the stages of a forecasting experiment (Figure~\ref{fig:pipeline}). The graph, the architecture and the training regime are configuration dictionaries consumed by fixed execution code, so one factor varies at a time without bespoke scripts. The four dictionaries of Figure~\ref{fig:pipeline} drive the whole load study of Section~\ref{sec:cases}.

\subsection{Data handling and graph construction}\label{sec:data}
The \code{DataClass} reads training and test tables, applies the temporal split, and manages calendar encodings and lagged features. The \code{GraphDataset} turns the tables into graph snapshots and fits feature and target scalers on the training window only. A single field of its configuration sets the forecasting horizon.

The \code{GraphBuilder} implements the adjacency constructions proposed for load forecasting \citep{campagne2025graph,campagne2026diagnostic}, each returning positive edge weights that grow with the strength of the relation between two nodes. A spatial graph thresholds a Gaussian kernel of geodesic distance. Data-driven graphs use dynamic time warping (DTW) \citep{salvador2004fastdtw}, correlation and precision matrices, a spline distance between temperature-to-load responses, or GL3SR, which learns a Laplacian from graph signals under a smoothness criterion \citep{humbert2021learning}. GraphToolbox can evaluate candidate graphs by training the same forecasting model on each and selecting the one with the lowest validation error. Graph construction precedes model training, so approaches that jointly learn the graph and the forecaster, such as TimeGNN \citep{xu2024timegnn}, are not covered by this procedure.

\subsection{Models and the convolution adapter}\label{sec:models}
One meta-model, \code{myGNN}, expresses every static architecture: a linear encoder lifts node features to a fixed latent width, residual message-passing blocks in the pre-activation DeepGCN arrangement \citep{li2019deepgcns} refine this latent representation, and a linear read-out returns the multi-step forecast at every node in one pass. Each block couples a convolution with layer normalization and a rectified linear activation, and multi-head convolutions split the latent width across heads so that the head count inflates neither capacity nor computation.

The user supplies the convolution itself by name. The meta-model receives a convolution class together with its keyword arguments and instantiates it through a \code{ConvAdapter}. The operators of PyG obey no common calling convention, some expecting scalar edge weights, others vector edge attributes, a relation type, or a positional input, and many expecting none of these. The adapter reads the constructor signature of the operator it is given, forwards only the arguments that operator accepts, and fills in the defaults it requires, whether a Chebyshev order, a relation count for the relational convolutions, or the small perceptrons that the isomorphism and edge convolutions expect. It then reads the forward signature, routes to the operator only the graph tensors it consumes, and supplies a neutral placeholder for any input the operator demands yet the forecasting graph does not carry. When an operator returns a width different from the latent one, as the multi-hop and signed convolutions do, a linear projection restores it. The few operators with CPU-only kernels run on the CPU, and the adapter moves their tensors between devices. The same path can expose the coefficients of the attention operators, which the interpretability module later renders as edge weights.

Swapping a graph convolutional network for an attention network or a Chebyshev filter is then the one-field change of Listing~\ref{lst:usage}. The adapter implements each operator-specific case once, in one place. Argument routing occupies about $240$ lines of adapter code, and the significance module a further $340$. The supported families include degree-normalized spectral convolutions \citep{kipf2017semisupervised}, inductive neighborhood aggregators \citep{hamilton2017inductive}, personalized propagation through APPNP \citep{gasteiger2018predict}, attention operators GATConv, GATv2Conv, and TransformerConv \citep{velivckovic2018graph,brody2021attentive,shi2020masked}, and polynomial spectral filters \citep{defferrard2016convolutional}.

The \code{ConvAdapterTemporal} adapter extends this interface to the recurrent graph cells of PyG-T \citep{rozemberczki2021pytorch}. It reads the constructor signature to distinguish a step-recurrent cell, which consumes one time step at a time while carrying a graph-aware hidden state, from a windowed cell, which ingests the whole window and must be told the number of periods in advance. Its companion model, \code{TemporalGNN}, replaces the static meta-model for these cells. It separates the ordered lag sequence of the target from the static side features, unrolls that sequence through the cell into a hidden state, concatenates the static features to that state, and projects the result to the forecast horizon. A numerically verified fast unroll of the Chebyshev gated cell folds the input-side gate transforms over the whole sequence and computes the graph normalization once, reverting to the reference step loop whenever an equivalence check fails. Since this model exposes the attributes that the training loop reads, a recurrent cell trains, reconciles, and caches exactly as a static convolution does, so the benchmark of Section~\ref{sec:cases} can place the two families under one budget.

\begin{lstlisting}[style=gtb, float=!htbp, floatplacement=!htbp, captionpos=b, abovecaptionskip=6pt, caption={Swapping a convolution is a one-field change. The same \texttt{Trainer} and \texttt{GraphDataset} serve any operator exposed through \texttt{ConvAdapter}.}, label={lst:usage}]
from torch_geometric.nn.conv import GATConv
from graphtoolbox.models import myGNN
from graphtoolbox.training import Trainer

model = myGNN(
    in_channels=ds_train.num_node_features,
    num_layers=2, hidden_channels=256,
    out_channels=48, # forecast horizon
    conv_class=GATConv, # <- swap operator here
    conv_kwargs={"heads": 2})

trainer = Trainer(model, ds_train, ds_val, ds_test,
                  batch_size=16, model_kwargs={"lr": 1e-3})
pred, target, edge_index, attn = trainer.train()
\end{lstlisting}

\begin{table}[!htbp]
\caption{Training-seed variability of the leading model of each study. National RMSE (MW), mean $\pm$ standard deviation over $5$ seeds at the configuration of the main tables, against the run reported in Tables~\ref{tab:load} and~\ref{tab:netload}, trained separately from the $5$.}
\label{tab:seeds}
\centering
\footnotesize
\begin{tabular}{@{}lcc@{}}
\toprule
\textbf{Task and model} & \textbf{5 seeds} & \textbf{Reported run} \\
\midrule
Load, GatedGraphConv & $927 \pm 31$ & $881$ \\
Net-load direct, FastRGCNConv & $1441 \pm 58$ & $1324$ \\
Net-load decomposed, FastRGCNConv & $1229 \pm 47$ & $1168$ \\
\bottomrule
\end{tabular}
\end{table}

\subsection{Training, optimization, and aggregation}\label{sec:training}
The \code{Trainer} manages the optimization loop, early stopping, and the evaluation of the standard forecasting metrics, namely mean absolute error, its normalized variant, MAPE, root mean squared error (RMSE), and bias. Optimization relies on Adam \citep{kingma2014adam}. A \code{RollingTrainer} extends this loop to a rolling-origin protocol, in which the training window expands to absorb each past test window before predicting the next. This protocol mirrors the retraining of a forecasting system in operation. An \code{Optimizer} built on Optuna \citep{akiba2019optuna} runs the hyperparameter search, so that the number of layers, the hidden width, the learning rate, the batch size, and even the graph become tunable through a declarative search space. A trial stops once its validation loss has not improved for a set number of epochs, and the pruner of the study abandons trials that fall behind the median of earlier ones. A seed fixes the sampler as well as the initialization and the batch order of every trial, and the trainer accepts the same seed, so a user can replay a search and the training runs it selects.

Because different architectures carry different inductive biases, their errors differ, and a combination can be more accurate than its best member. The framework provides a native \code{Aggregation} module that performs sequential expert aggregation either online or in batch, following the aggregation rules of the opera package \citep{gaillard2021opera}, whose R and Python\footnote{\url{https://github.com/Dralliag/opera-python}} reference implementations informed our design. Its default rule is the parameter-free MLpol algorithm \citep{gaillard2014second}, which maintains a weight per expert and updates it according to past forecasting performance, so that the aggregated prediction is a time-varying convex combination of the experts. The exponentially weighted average forecaster \citep{vovk1990aggregating,cesa2006prediction} and Bernstein online aggregation \citep{wintenberger2017optimal} are also available. A uniform average serves as a parameter-free reference, and an ensemble of independently initialized models supplies a first estimate of predictive spread. Forecasts can be aggregated across architectures or across random initializations of one architecture, either at the national level (top-level) or separately for each node (bottom-up).

\begin{figure}[!htbp]
\centering
\includegraphics[width=\figcol]{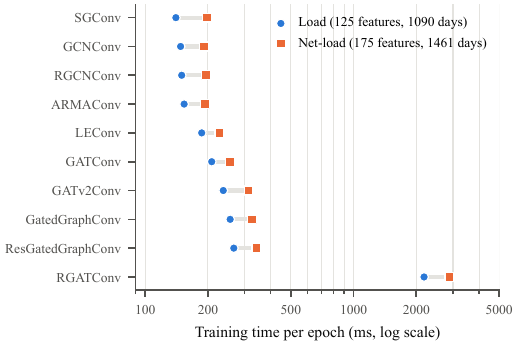}
\caption{Training time per epoch for the $10$ convolutions in Table~\ref{tab:load}, evaluated on load and net-load with one layer, hidden width $64$, and the DTW graph. Times are averaged over $20$ epochs on an Apple M4 Pro CPU. The load runs use $125$ input features, against $187$ in the load sweep.}
\label{fig:timing}
\end{figure}

\subsection{Interpretability and visualization}\label{sec:interp}
The interpretability module provides feature- and edge-level diagnostics for individual forecasts. Accumulated local effects estimate the marginal influence of a feature while accounting for its correlation with other inputs, which makes them more reliable than partial dependence in the highly collinear setting of weather and calendar drivers \citep{apley2020visualizing}. For a graph model, accumulated local effects are estimated by varying one feature at one node across quantile bins while keeping other inputs at their observed values. Changes are aggregated over all nodes, thereby including the effects propagated through message passing. The framework also visualizes edge attributions obtained from a user-run PyG explainer, from attention weights when available, or from occlusion. The latter measures how much the forecast at one node changes when another node's inputs are replaced by a baseline \citep{ying2019gnnexplainer}. These diagnostics are correlational, and we make no claim that they establish a causal link between a connection and a prediction. They are also bounded by the attribution they are given, and Section~\ref{sec:cases} shows attention weights that reduce to one score per sending region on one graph and, on another, agree only moderately with the effect of a neighbor on the forecast. Beyond these post-hoc attributions, the package also includes an intrinsically interpretable model, an additive graph network that encodes each feature group with its own subnetwork and sums their contributions, so that a prediction decomposes into per-group terms, each of which can be read on its own. A visualization module places node errors, learned graphs, and edge weights on maps of the corresponding territory.

\subsection{Significance testing}\label{sec:signif}
An accuracy ranking without a measure of its reliability leaves open whether a gap reflects a real difference or the noise of a finite test window. The evaluation module operates entirely on cached forecasts, allowing the statistical analysis to run on stored predictions without retraining the models. The Diebold--Mariano test \citep{diebold1995comparing} assesses pairwise predictive accuracy under a squared, absolute, or percentage loss, with a Newey-West long-run variance \citep{newey1987simple} and the small-sample correction of Harvey, Leybourne, and Newbold \citep{harvey1997testing}. A Holm-Bonferroni adjustment controls the family-wise error rate across the pairwise comparisons. A moving-block bootstrap \citep{kunsch1989jackknife} quantifies the sampling uncertainty on a metric, with a default block of one day and $2000$ resamples. Its blocks of consecutive steps preserve the autocorrelation of forecast errors that an ordinary resampling would destroy. To summarize an entire sweep, the module reports the Model Confidence Set \citep{hansen2011model}, the subset of architectures that contains the best one with a prescribed probability.

These instruments address the sampling uncertainty of a metric on a finite test window, and they leave a second source untouched, the stochasticity of training itself. Because the sweeps of Section~\ref{sec:cases} report one run per architecture, we retrained the strongest operators over $5$ seeds at their table configurations. Table~\ref{tab:seeds} sets the resulting means against the entries of the main tables, which sit $46$ to $117$~MW below them. Two conclusions remain stable across seeds. At the seed mean, the graph models stay below the recurrent baseline, by $151$~MW on load and by $305$ and $517$~MW for the direct and decomposed net-load models, which also stay below the regional GAM. For net-load, the paired improvement from direct to decomposed forecasting averages $212$~MW with a standard deviation of $81$~MW. The fine ordering within each architectural band is not stable, which is why we claim no unique best convolution.

\subsection{Scale}
On the French graph with $12$ nodes, $9$ of the $10$ representative convolutions require between $140$ and $270$~ms per training epoch on load and between $190$ and $340$~ms on net-load (Figure~\ref{fig:timing}), corresponding to approximately one minute for a full training run. Only the relational-attention RGATConv is an order of magnitude slower. We measured these timings on the CPU of a MacBook Pro with an Apple M4 Pro ($12$ cores), which outperforms the Metal Performance Shaders (MPS) backend at this graph size.

We repeat the measurement using the same code and hardware on WEAVE-UK, a dataset of half-hourly consumption from $500$ low-voltage feeders operated by Scottish and Southern Electricity Networks. The observations cover 14--28 February 2024 and were released through the Weave data platform\footnote{\url{https://weave.energy/}}~\weavecite. Inference per forecast window rises from under a millisecond to $14$~ms for GraphSAGE, $40$~ms for GAT and $71$~ms for APPNP (Table~\ref{tab:footprint}). The correlation graph used for WEAVE in the absence of coordinates connects almost every pair of feeders, with $242{,}624$ edges against $142$ on the French graph, so the $40$-fold increase in node count multiplies the number of edges by about $1700$ and the latency by $47$ to $144$. Accuracy results for graph and non-graph models on these $500$ feeders are reported in \citet{campagne2026diagnostic}. These measurements show that sweeps on graphs of this size remain feasible on a workstation. They do not characterize computational cost on graphs with thousands of nodes.

\begin{table}[!htbp]
\caption{Cost of the same code path at two graph scales, on the $12$ French regions and on $500$ WEAVE-UK feeders, measured on the Apple M4 Pro CPU at hidden width~$64$ and $2$ layers. Parameter counts differ between the two columns because the input dimension does ($187$ features against $50$). Inference time is the mean over $200$ calls on one test window.}
\label{tab:footprint}
\centering
\footnotesize
\begin{tabular}{@{}lcccc@{}}
\toprule
& \multicolumn{2}{c}{\textbf{12 nodes}} & \multicolumn{2}{c}{\textbf{500 nodes}} \\
\cmidrule(lr){2-3}\cmidrule(lr){4-5}
\textbf{Operator} & params (k) & infer (ms) & params (k) & infer (ms) \\
\midrule
GraphSAGE & 32.0 & 0.30 & 23.3 & 14.0 \\
GAT & 24.4 & 0.44 & 15.6 & 40.2 \\
APPNP & 15.5 & 0.49 & 6.8 & 70.7 \\
\bottomrule
\end{tabular}
\end{table}

\section{Convolution coverage}\label{sec:coverage}
We measure the breadth of the framework by the fraction of available operators it can run. We instantiate each of the $65$ convolutions exported by \code{torch\_geometric.nn.conv} in PyG 2.6.1 inside the two-layer meta-model and train it for one step on a complete homogeneous graph with random node features and edge weights. Table~\ref{tab:coverage} reports the outcome by family. Of the $65$ tested operators, $53$ ($81.5\%$) run without modification. A single operator, the fused attention kernel, requires an optional dependency that is unavailable on some platforms. The remaining $11$ presuppose heterogeneous graphs, hypergraphs, point-cloud inputs, or device-specific libraries, and therefore fall outside the homogeneous node-level regime that electricity forecasting occupies. This coverage test measures whether an operator instantiates and completes a training pass, and does not by itself certify forecasting accuracy. Of the $53$ operators supported by the adapter, $46$ enter the load-forecasting sweep and its significance analysis. We exclude $7$ operators from the forecasting sweep: DNAConv, GPSConv, PointTransformerConv, SplineConv, WLConv, WLConvContinuous, and XConv. We report a representative subset, including the $10$ static convolutions in Table~\ref{tab:load} and the recurrent GConvGRU and DCRNN cells. Table~\ref{tab:operators} summarizes the neighbor-aggregation rules of the operators discussed in Section~\ref{sec:cases}. Only $8$ use edge weights directly; the others receive information from the dynamic-time-warping graph only through its connectivity.

\begin{table}[!htbp]
\caption{Coverage of PyG convolutional operators within the \code{myGNN} meta-model. Operator families follow Table~\ref{tab:operators}, with one additional family for operators not discussed in Section~\ref{sec:cases}. Skipped operators require heterogeneous graphs, hypergraphs, point-cloud inputs, or device-specific libraries.}
\label{tab:coverage}
\centering
\footnotesize
\begin{tabularx}{0.8\textwidth}{@{}l c >{\raggedright\arraybackslash}X@{}}
\toprule
\textbf{Operator family} & \textbf{Working} & \textbf{Examples} \\
\midrule
Propagation & 13 & GCN, SG, APPNP \\
Attention & 10 & GAT, GATv2 \\
Gating and modulation & 3 & GatedGraph, FiLM \\
Relational and edge-conditioned & 9 & RGCN, NN, GMM \\
Point-cloud and dynamic graphs & 4 & Edge, GravNet \\
Other aggregators & 14 & SAGE, GIN, PNA \\
\midrule
\textbf{Working total} & \textbf{53} & (81.5\%) \\
Optional dependency & 1 & FusedGAT \\
Skipped (out of scope) & 11 & see caption \\
\textbf{Tested total} & \textbf{65} & \\
\bottomrule
\end{tabularx}
\end{table}

\begin{table}[!htbp]
\caption{Operators named in Section~\ref{sec:cases}, grouped by aggregation mechanism. Each operator is used within the residual block of the meta-model. A check mark indicates whether it uses edge weights. The graphs considered here have a single edge type and no edge attributes. Relational operators therefore use one relation, while GMMConv receives constant edge coordinates.}
\label{tab:operators}
\centering
\footnotesize
\begin{tabularx}{0.8\textwidth}{@{}l@{\hspace{3pt}}r >{\raggedright\arraybackslash}X c@{}}
\toprule
\textbf{Operator} & & \textbf{Aggregation of the neighbors} & \textbf{Wt.} \\
\midrule
\rowcolor{black!8}\multicolumn{4}{@{}l}{\textit{Propagation}} \\
GCNConv & \citep{kipf2017semisupervised} & degree-normalized sum & \checkmark \\
SGConv & \citep{wu2019simplifying} & $K$ steps, then one linear map & \checkmark \\
SSGConv & \citep{zhu2021simple} & mean of $K$ propagation steps & \checkmark \\
APPNP & \citep{gasteiger2018predict} & personalized-PageRank propagation & \checkmark \\
GCN2Conv & \citep{chen2020simple} & propagation with initial residual & \checkmark \\
ARMAConv & \citep{bianchi2021graph} & rational spectral filter & \checkmark \\
PANConv & \citep{ma2020path} & sum over paths up to a set length & \\
\rowcolor{black!8}\multicolumn{4}{@{}l}{\textit{Attention}} \\
GATConv & \citep{velivckovic2018graph} & attention ranked by the sender & \\
GATv2Conv & \citep{brody2021attentive} & attention ranked per receiver & \\
SuperGATConv & \citep{kim2021how} & attention with edge self-supervision & \\
RGATConv & \citep{busbridge2019relational} & attention per relation type & \\
\rowcolor{black!8}\multicolumn{4}{@{}l}{\textit{Gating and modulation}} \\
GatedGraphConv & \citep{li2016gated} & GRU update from summed messages & \checkmark \\
ResGatedGraphConv & \citep{bresson2017residual} & messages scaled by edge gates & \\
FiLMConv & \citep{brockschmidt2020gnn} & messages modulated by the receiver & \\
\rowcolor{black!8}\multicolumn{4}{@{}l}{\textit{Relational and edge-conditioned}} \\
(Fast)RGCNConv & \citep{schlichtkrull2018modeling} & one weight matrix per relation type & \\
GMMConv & \citep{monti2017geometric} & Gaussian kernels of edge coordinates & \\
FeaStConv & \citep{verma2018feastnet} & soft assignment to weight matrices & \\
\rowcolor{black!8}\multicolumn{4}{@{}l}{\textit{Other aggregators}} \\
LEConv & \citep{ranjan2020asap} & weighted node--neighbor differences & \checkmark \\
PNAConv & \citep{corso2020principal} & mean, maximum and sum combined & \\
GENConv & \citep{li2023deepergcn} & softmax-weighted mean & \\
\bottomrule
\end{tabularx}
\end{table}

\section{Controlled benchmark}\label{sec:cases}
Each comparison in this section changes a single choice of one pipeline configuration: the convolution among the $46$ operators of the sweep, the graph against the identity graph, the aggregation rule, the reconciliation, recurrent against static cells, and, on net-load, the weather information and the decomposition of the target. Within each task, apart from the hyperparameter search on $3$ operators, the data, the 2019 test year, the graph and the architecture stay fixed, so that we can attribute a difference between two rows to the choice that separates them. Both targets are forecast for the $12$ French regions and scored on their national sum. Writing $y_t$ for the national target at half-hour~$t$ of the test year, $\hat y_t$ for its forecast and $T$ for the number of half-hours, the tables report
\begin{equation}
\mathrm{RMSE} = \Big(\frac{1}{T}\sum_{t=1}^{T}(\hat y_t-y_t)^2\Big)^{1/2},\quad
\mathrm{MAPE} = \frac{100}{T}\sum_{t=1}^{T}\frac{|\hat y_t-y_t|}{|y_t|},\quad
\mathrm{sMAPE} = \frac{100}{T}\sum_{t=1}^{T}\frac{2\,|\hat y_t-y_t|}{|\hat y_t|+|y_t|}.
\end{equation}
RMSE is expressed in megawatts, and the rows within each sweep are ordered by this metric. We also report MAPE for load and sMAPE for net-load. The latter is bounded by $200\%$ and remains defined near zero target values. We assess differences using two-sided paired Diebold--Mariano tests on squared errors over the test year, with a significance threshold of $p<10^{-5}$. Load is positive and relatively smooth, whereas net-load can change sign and reflects fluctuations in wind and solar production.

The references, except for persistence and the two foundation models, are fitted once on the training years, like the graph models. Persistence repeats the value of the previous day at the same half-hour. Chronos-Bolt and Chronos-2, two time-series foundation models of the Chronos family \citep{ansari2024chronos}, forecast without training on these data. ARIMA-X fits one seasonal autoregressive integrated moving average model with exogenous covariates per half-hour, and XGBoost \citep{chen2016xgboost} fits one model per half-hour of the horizon, either on the samples of all regions pooled or per region. A multilayer perceptron (MLP) is fitted per region, and a long short-term memory (LSTM) network \citep{hochreiter1997long} is conditioned on the covariates. The additive references fit one model per half-hour on weather and calendar covariates, nationally on the aggregate or per region with the regional forecasts summed. Besides the GAMs, TAM \citep{tam2026package} fits the regional version with a temperature-weekend tensor and one smoothing parameter per effect selected by generalized cross-validation, and reads no lagged target. XGBoost and the MLP read the inputs of the graph models and follow the search spaces and seeds of \citet{campagne2026diagnostic}, which score $10$ configurations on 2018.

\begin{figure*}[!htbp]
\centering
\includegraphics[width=\textwidth]{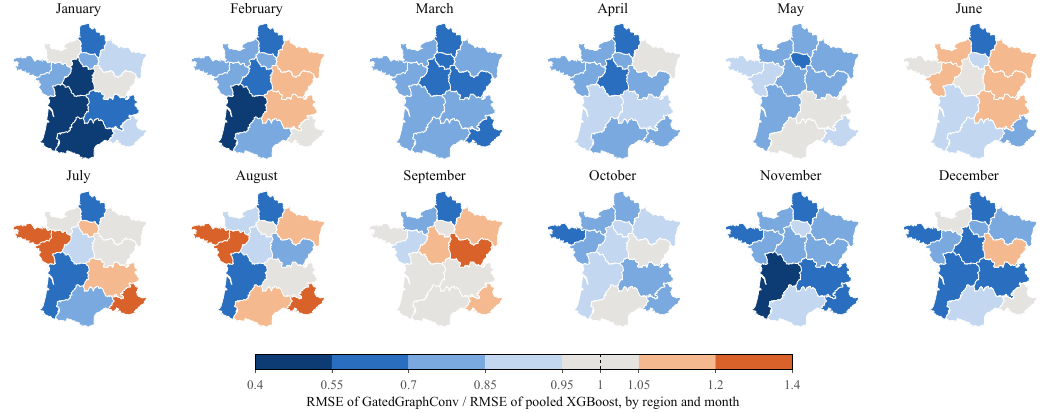}
\caption{GatedGraphConv against pooled XGBoost on the $12$ French regions, 2019 test set: RMSE of GatedGraphConv divided by that of XGBoost (seed $42$) over the half-hours of each month, in blue where the graph model is more accurate.}
\label{fig:regionalerror}
\end{figure*}

\subsection{Load forecasting}
The first task forecasts French electricity load at the granularity of the $12$ administrative regions supplied by the transmission operator, using half-hourly RTE open data \citep{rte_ecowatt} enriched with calendar features and M\'et\'eo-France SYNOP weather observations \citep{meteofrance_synop}. The nodes of the graph are the regions, each positioned at the spatial coordinates of a major city of that region. A model reads its features at midnight and returns load forecasts for the $48$ half-hours of the day that follows. The $48$ half-hourly load lags predate that origin, whereas the weather of the forecast day enters as its observed values. Every model in the comparison that uses weather shares this perfect-forecast assumption, which makes all errors lower than those of an operational forecast built on numerical weather predictions. The sweep holds the graph at the dynamic-time-warping (DTW) construction and the architecture at $2$ layers of width $64$ for every operator, so every row of Table~\ref{tab:load} comes from the same implementation, configured once per operator. We run the optimizer of the framework separately on $3$ operators, searching their model and graph over $100$ trials each, which we score on the 2018 validation year. The search selects the spatial graph for APPNP, the spline-distance graph for LEConv, and the identity graph for GATConv, which removes message passing between regions. On the 2019 test set, it lowers the RMSE of APPNP from $1093$ to $971$~MW and that of GATConv from $923$ to $896$~MW, whereas LEConv does not improve ($932$ against $939$~MW), so a configuration selected on one validation year does not always transfer to the next.

Table~\ref{tab:load} collects the benchmark on the 2019 test set. Across the $46$ evaluated convolutions, MAPE ranges from $1.20\%$ to $1.60\%$, with a median of $1.31\%$. GatedGraphConv has the lowest RMSE at $881$~MW, and GCNConv the lowest MAPE at $1.20\%$.

Every operator in Table~\ref{tab:load} improves on the additive, recurrent and foundation references: the national and regional additive models reach $1228$ and $1200$~MW, the LSTM $1078$~MW, Chronos-2 $1647$~MW, and TAM $1062$~MW. The gap of TAM to the graph models ranges from $130$~MW at the weakest listed operator to $181$~MW at the best. The second gap is significant, whereas the first is not ($p=1.5\times10^{-4}$, against a threshold of $10^{-5}$). XGBoost given the same inputs as the graph models is more accurate. Pooled over the regions, it reaches $805$~MW in Table~\ref{tab:load} (its seed-$42$ run, used below, reaches $804$~MW), $76$~MW below the best operator, a difference that is not significant, and $122$~MW below the mean of the $5$ seeds of GatedGraphConv (Table~\ref{tab:seeds}). Averaging the forecasts of these $5$ seeds lowers the RMSE of GatedGraphConv to $824$~MW, which does not differ significantly from pooled XGBoost. Fitted per region, it reaches $872$~MW, and the MLP fitted per region $944$~MW, within the band of the sweep. Region by region, GatedGraphConv is the more accurate of the two (Figure~\ref{fig:regionalerror}). Its RMSE over the test year is lower in each of the $12$ regions, significantly in $10$ of them, and in $114$ of the $144$ region-months, $24$ of the $30$ exceptions falling between June and September. Its national RMSE is nevertheless higher, because its regional errors compensate less when summed. Were these errors uncorrelated, the national RMSE would be $458$~MW for GatedGraphConv against $589$~MW for XGBoost (seed $42$), and their correlation, $0.27$ on average against $0.16$, raises it to $881$ and $804$~MW. Averaging $5$ seeds leaves the first quantity at $458$~MW and lowers the national RMSE to $824$~MW.

This comparison alone cannot establish the benefit of spatial message passing, since the graph models also differ from the baselines in their inputs, parameter sharing across nodes, and multi-step prediction architecture. Isolating message passing requires holding everything else fixed. We retrain the same $10$ operators in dedicated paired runs at a strictly common configuration on the DTW and identity graphs, so these runs are not directly comparable to the entries of Table~\ref{tab:load}. Averaging their individual RMSEs gives $951$~MW on the DTW graph and $984$~MW on the identity graph. The median paired reduction is $23$~MW, and $8$ of the $10$ operators improve. The effect is architecture-dependent, with the identity graph improving $2$ operators. Message passing thus contributes to the improvement over the references, although it does not explain the entire gap. A companion study using GraphToolbox separates the effects of parameter sharing and neighbor information by comparing graph models with independent per-node networks and a shared network without message passing \citep{campagne2026diagnostic}. On $500$ low-voltage feeders, parameter sharing alone explains $78$ to $84\%$ of the performance gap between the independent per-node model and the graph model.

\begin{figure}[!htbp]
\centering
\includegraphics[width=\textwidth]{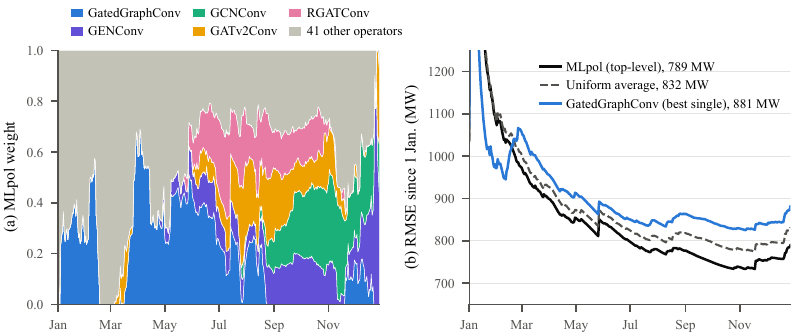}
\caption{Online MLpol aggregation of the $46$ load-forecasting operators on the 2019 test year. (a) Daily weights of the $5$ operators with the largest average weights. (b) Cumulative RMSE since 1 January for MLpol, uniform averaging, and the best individual operator. The first days are omitted for readability.}
\label{fig:aggregation}
\end{figure}

\begin{figure}[!htbp]
\centering
\includegraphics[width=\figcol]{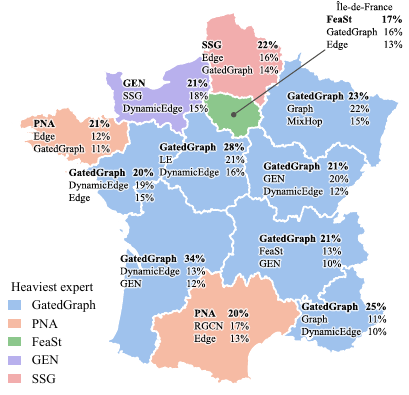}
\caption{Bottom-up MLpol aggregation across the $12$ French regions in 2019. For each region, the $3$ experts with the highest mean weights are shown, with the largest in bold. The map color indicates the highest-weighted expert. Operator names omit the suffix Conv.}
\label{fig:regional}
\end{figure}

\begin{figure*}[!htbp]
\centering
\includegraphics[width=\textwidth]{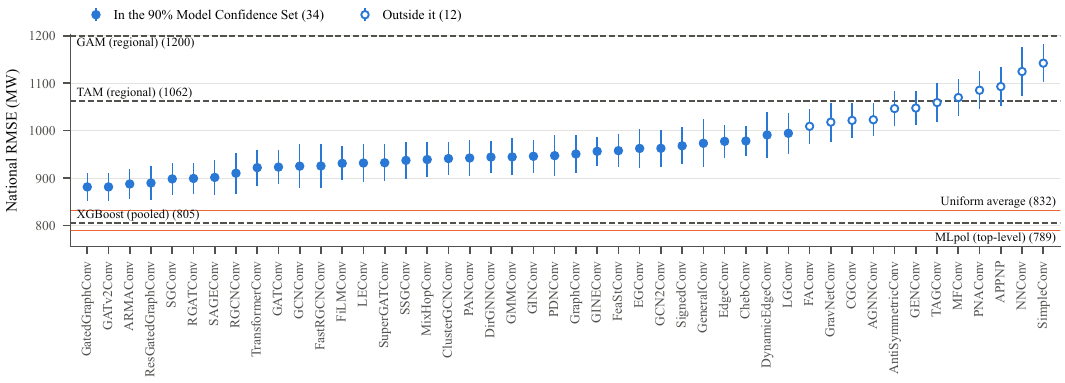}
\caption{National RMSE of the $46$ operators of the load sweep on the 2019 test set, with block-bootstrap standard errors. Filled markers belong to the $90\%$ Model Confidence Set. Solid lines mark the two aggregates of the sweep, dashed lines $3$ references of Table~\ref{tab:load}.}
\label{fig:sweep}
\end{figure*}

Online expert aggregation over the whole sweep lowers the MAPE from $1.11\%$ for a uniform average to $1.03\%$ for the MLpol mixture, at a national RMSE of $789$~MW, $92$~MW below the best single convolution. MLpol sets its weights each day from the losses of the previous test days, so it forecasts online, whereas the uniform average, like weights fixed on the validation year, stays offline. The mixture and pooled XGBoost do not differ significantly. The block-bootstrap intervals of Table~\ref{tab:load} and Figure~\ref{fig:sweep} do not resolve the ordering inside the sweep, and neither do statistical tests: a Model Confidence Set at the $90\%$ level, computed on the same cached forecasts, retains $34$ of the $46$ operators, including the $10$ listed in the table, and a Holm-corrected Diebold--Mariano test separates none of those $10$ from the best. The mixture is nevertheless significantly more accurate than the best single convolution, despite the overlap of their marginal bootstrap intervals. The mixture moves its weight across operators during the year (Figure~\ref{fig:aggregation}). From January to April, GatedGraphConv carries $27\%$ of the weight on average, and from June to October RGATConv and GATv2Conv carry $18\%$ and $16\%$. The cumulative error of the mixture stays below that of the best single operator from 18 February onward. Aggregation therefore removes the need to identify the best operator in advance, although it still requires running the whole sweep. Run region by region, the bottom-up mixture selects different experts (Figure~\ref{fig:regional}). GatedGraphConv carries the largest weight in $7$ of the $12$ regions, PNAConv leads in Bretagne and Occitanie, and in every region the mixture is more accurate than the best single operator of that region, by $1.9\%$ to $9.7\%$ of that operator's RMSE.

As an alternative to combining fixed models, GraphToolbox can update a single model during forecasting. Each day, the rolling trainer produces a forecast and then performs a small number of Adam steps using mini-batches drawn partly from the preceding $7$ days and partly from the full training history. The graph and feature scalers remain fixed. We initialize the procedure from the $5$ GatedGraphConv checkpoints in Table~\ref{tab:seeds}, select the learning rate ($3\times10^{-5}$) and daily update count ($30$) on 2018, and run the updates through 2018 before evaluating on 2019. This aligns the historical information available to the graph model with that of the refitted TAM. Daily updates reduce mean RMSE from $927 \pm 31$ to $752 \pm 34$~MW. The regional TAM adapted each day through a Kalman filter, following \citet{obst2021adaptive} with its two variances tuned on 2018, reaches $751$~MW. None of the $5$ updated GatedGraphConv runs differs significantly from this reference. Averaging their forecasts further reduces RMSE to $678$~MW.

Minimum-trace (MinT) reconciliation \citep{wickramasuriya2019optimal} combines regional predictions with a national forecast while enforcing coherence between them. GraphToolbox estimates reconciliation weights from the 2018 validation errors, using either their variances or their full covariance shrunk toward its diagonal. The national forecast can be generated within the framework or supplied externally. An XGBoost model fitted on the aggregate series using lagged load, calendar information, and national means of the forecast-day covariates (including weather) reaches $1198$~MW. Reconciling GatedGraphConv with this forecast lowers RMSE from $881$ to $844$~MW, a significant improvement. Using instead the national sum of pooled XGBoost, the strongest reference in Table~\ref{tab:load}, whose seed-$42$ run reaches $804$~MW, gives $802$~MW with diagonal covariance and $747$~MW with full covariance estimated by the shrinkage method of \citet{schafer2005shrinkage}. Both improve significantly on the original GatedGraphConv forecast, but neither differs significantly from the national XGBoost forecast. Reconciliation with TAM, the strongest additive baseline, yields $818$ and $784$~MW with diagonal and full covariance, respectively, also significant improvements. TAM uses forecast-day calendar and weather information without load lags, whereas pooled XGBoost uses the graph model's inputs. Reconciliation therefore does not introduce information beyond the common day-ahead forecasting setting. The final blocks of Tables~\ref{tab:load} and~\ref{tab:netload} summarize these results. 

\begin{table}[!htbp]
\caption{French national load, 2019 test set. MAPE and RMSE (MW) for $10$ of the $46$ operators of the load sweep, which are not the $10$ most accurate (Figure~\ref{fig:sweep} reports all $46$), ordered by RMSE. Forecasts are not reconciled except in the final block, where MinT combines GatedGraphConv regional forecasts with the national forecast from pooled XGBoost, the strongest reference, using diagonal or full error covariance estimated on 2018. Uncertainties for GNN rows are block-bootstrap standard errors from one training run (see Table~\ref{tab:seeds}). Baseline results are point estimates, averaged over $3$ seeds for pooled XGBoost and $10$ for the MLP. The reproducibility section identifies results reused from previous studies and those obtained for this paper.}
\label{tab:load}
\centering
\footnotesize
\setlength{\tabcolsep}{4pt}
\begin{tabular}{@{}w{l}{94.5pt}w{c}{52pt}w{c}{52pt}@{}p{9pt}@{}w{l}{94.5pt}w{c}{52pt}w{c}{52pt}@{}}
\cmidrule[\heavyrulewidth]{1-3}\cmidrule[\heavyrulewidth]{5-7}
\textbf{Model} & \textbf{MAPE (\%)} & \textbf{RMSE (MW)} & & \textbf{Model} & \textbf{MAPE (\%)} & \textbf{RMSE (MW)} \\
\cmidrule[\lightrulewidth]{1-3}\cmidrule[\lightrulewidth]{5-7}
\multicolumn{3}{@{}>{\columncolor{black!8}[0pt][0pt]}l@{}}{\textit{Baselines}} & & \multicolumn{3}{@{}>{\columncolor{black!8}[0pt][0pt]}l@{}}{\textit{Static GNN sweep (GraphToolbox)}} \\
Persistence (1 day) & 5.78 & 4507 & & GatedGraphConv & $1.22 \pm 0.03$ & $881 \pm 29$ \\
Chronos-Bolt & 2.99 & 2408 & & GATv2Conv & $1.22 \pm 0.03$ & $881 \pm 30$ \\
ARIMA-X & 2.51 & 1706 & & ARMAConv & $1.24 \pm 0.03$ & $887 \pm 32$ \\
Chronos-2 & 1.78 & 1647 & & ResGatedGraphConv & $1.22 \pm 0.03$ & $890 \pm 36$ \\
GAM (national) & 1.69 & 1228 & & SGConv & $1.24 \pm 0.03$ & $898 \pm 33$ \\
GAM (regional) & 1.67 & 1200 & & RGATConv & $1.22 \pm 0.03$ & $899 \pm 34$ \\
LSTM & 1.46 & 1078 & & RGCNConv & $1.23 \pm 0.03$ & $910 \pm 43$ \\
TAM (regional) & 1.59 & 1062 & & GATConv & $1.25 \pm 0.03$ & $923 \pm 36$ \\
MLP (regional) & 1.31 & 944 & & GCNConv & $1.20 \pm 0.03$ & $925 \pm 46$ \\
XGBoost (regional) & 1.15 & 872 & & LEConv & $1.30 \pm 0.03$ & $932 \pm 39$ \\
XGBoost (pooled) & 1.11 & 805 & & & & \\
\cmidrule[\lightrulewidth]{1-3}\cmidrule[\lightrulewidth]{5-7}
\multicolumn{3}{@{}>{\columncolor{black!8}[0pt][0pt]}l@{}}{\textit{Expert aggregation}} & & \multicolumn{3}{@{}>{\columncolor{black!8}[0pt][0pt]}l@{}}{\textit{GatedGraphConv reconciled by MinT onto XGBoost (pooled)}} \\
Uniform average & $1.11 \pm 0.03$ & $832 \pm 38$ & & Diagonal covariance & $1.11 \pm 0.03$ & $802 \pm 26$ \\
MLpol (bottom-up) & $1.06 \pm 0.03$ & $803 \pm 38$ & & Full covariance & $1.03 \pm 0.03$ & $\mathbf{747 \pm 25}$ \\
MLpol (top-level) & $\mathbf{1.03 \pm 0.03}$ & $789 \pm 38$ & & & & \\
\cmidrule[\heavyrulewidth]{1-3}\cmidrule[\heavyrulewidth]{5-7}
\end{tabular}
\end{table}

The framework also runs T-GCN, A3T-GCN, DCRNN, and GConvGRU \citep{zhao2019tgcn,bai2020a3tgcn,li2018diffusion,seo2018structured}, either one-shot, updating once on the stacked lag channels as a static convolution does, or unrolled over the ordered lags. With one-shot forecasting, GConvGRU and DCRNN achieve RMSEs of $954$ and $1002$~MW, respectively. GConvGRU sits at the median of the static convolution sweep ($949$~MW) and DCRNN $53$~MW above it, and both are less accurate than its best operator ($881$~MW).

\subsection{Net-load forecasting}
\begin{figure*}[!htbp]
\centering
\includegraphics[width=\textwidth]{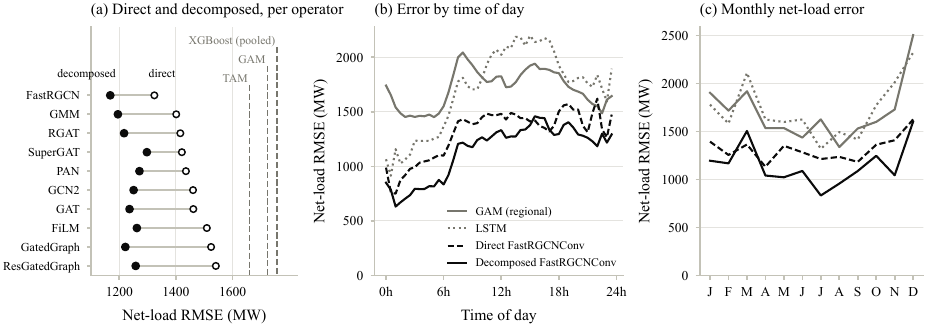}
\caption{Net-load forecasts on the 2019 test year. (a) National RMSE of direct (open markers) and decomposed (filled markers) forecasts for the $10$ operators in Table~\ref{tab:netload}, compared with pooled XGBoost ($3$-seed mean), TAM, and the regional GAM. (b) National RMSE by half-hour for the best direct and decomposed models (FastRGCNConv), regional GAM, and LSTM. (c) Monthly national RMSE for the same four forecasts, with line styles as in panel~(b). LSTM results are averaged over $5$ seeds.}
\label{fig:netload}
\end{figure*}

The second task forecasts national net-load, the demand that remains once solar and wind generation are subtracted, which is the quantity a system operator must balance. This target is harder than gross load, because it changes sign and inherits the volatility of renewable production. The percentage error is meaningless for series that pass through zero, such as solar generation at night, so the pipeline restricts that metric to strictly positive series and reports RMSE and a symmetric percentage error elsewhere.

Each net-load model receives, besides the net-load lags of the previous day, the trajectories of temperature, wind speed, and cloud cover over the $48$ half-hours of the forecast day, with the power-weighted versions of wind and cloud cover used by the additive references. On the same 2019 test set (Table~\ref{tab:netload}), the $10$ convolutions, all at the same validation-selected configuration ($1$ layer of width $64$), occupy a band from $2.19\%$ to $2.52\%$ sMAPE, with FastRGCNConv achieving the lowest RMSE at $1324$~MW, and the uniform average of the direct sweep reaches $1220$~MW.

Persistence, the zero-shot Chronos-Bolt at $3676$~MW, and ARIMA-X at $2933$~MW trail far behind, and Chronos-2 reaches $2344$~MW. Among the fitted baselines, pooled and region-specific XGBoost achieve $1757$ and $1938$~MW, respectively. The national GAM reaches $1763$~MW, the LSTM $1746$~MW, the regional GAM $1723$~MW, regional TAM $1659$~MW, and the regional MLP $1473$~MW. All direct convolutions outperform the additive, recurrent, and XGBoost references, with a margin of at least $118$~MW. The best convolution improves on pooled XGBoost, which uses the same inputs, by $433$~MW. The MLP lies within the direct convolution range. Unlike the graph models, the additive references use weather covariates at each forecast half-hour without lagged net-load. Without the wind and cloud-cover trajectories, the best direct graph model reaches $2094$~MW, above every fitted baseline, so its advantage over the additive references requires these inputs.

Net-load also decomposes into physical components, $\text{net-load} = \text{load} - \text{wind} - \text{solar}$. Forecasting each component with its own GNN and recombining the forecasts, using the same configuration and training budget per model, improves the forecasts of all $10$ convolutions and lowers RMSE by $211$~MW on average. FastRGCNConv reaches $1.91\%$ and $1168$~MW, and the uniform average of the decomposed forecasts $1.78\%$ and $1098$~MW. The best decomposed forecast is significantly more accurate than the best direct one. It is $491$~MW below TAM, $555$~MW below the regional GAM and $578$~MW below the recurrent model, and all three differences are significant (for the recurrent model, against each of the $5$ LSTM seeds). The same decomposition lowers the RMSE of pooled XGBoost to $1678$~MW, still $510$~MW above the decomposed FastRGCNConv, and every decomposed graph model is more accurate than the MLP. The decomposition also costs three models for one forecast, so the comparison is not matched on total training compute.

The decomposition narrows the spread of the $10$ operators, from $1324$--$1541$~MW for the direct forecasts to $1168$--$1297$~MW (Figure~\ref{fig:netload}a). Across operators, the RMSE of the wind component ranges from $786$ to $853$~MW, against $796$ to $904$~MW for load and $342$ to $428$~MW for solar, and the squared wind RMSE makes up $40\%$ to $45\%$ of the sum of the squared component RMSEs, which approximates its share of the recombined error when the component errors are independent. Over the day, the error of the best decomposed forecast concentrates in the half-hours between 7:00 and 20:00, which carry $67\%$ of its squared error, and that forecast has a lower RMSE than both references at each of the $48$ half-hours (Figure~\ref{fig:netload}b). Averaged over these daytime half-hours, the recurrent model has the largest error, $1964$~MW against $1849$~MW for the regional GAM and $1298$~MW for the decomposed FastRGCNConv, although it is ahead of the GAM during the rest of the day. The advantage over the references holds throughout the year: in each of the $12$ months, the best direct and decomposed forecasts have a lower RMSE than both the regional GAM and the recurrent model, and the decomposition improves the direct forecast in every month except March (Figure~\ref{fig:netload}c).

Reconciliation onto the national TAM forecast with the same diagonal weights lowers the RMSE of the direct FastRGCNConv forecast from $1324$ to $1238$~MW, a significant gain, whereas that of the decomposed forecast moves from $1168$ to $1162$~MW, a difference that is not significant. Reconciliation with the full error covariance gives $1236$ and $1160$~MW. With the national sum of the MLP fitted per region, the most accurate reference at $1398$~MW for the seed used here, reconciliation gives $1269$~MW for the direct forecast with either covariance, a gain that is significant with the diagonal one, and $1152$ and $1151$~MW for the decomposed forecast, differences that are not significant.

Unrolled over the $48$ lags and given the same weather trajectories, GConvGRU reaches $1848$~MW, above every direct convolution of Table~\ref{tab:netload} and above the regional GAM, so the static sweep keeps the advantage on net-load.

\begin{table}[!htbp]
\caption{French national net-load, 2019 test set. sMAPE and RMSE (MW) for direct and decomposed GNNs and their aggregations. Direct and decomposed models use the same per-model configuration and training budget. All models use observed weather from the target day. Uncertainties are block-bootstrap standard errors computed from a single training run (see Table~\ref{tab:seeds}). Baseline point estimates are averaged over $3$ seeds for pooled XGBoost, $5$ for the LSTM and $10$ for the MLP. Best values are in bold. The last block reconciles FastRGCNConv by MinT with the national sum of the MLP fitted per region (seed $42$), the most accurate reference, using the diagonal or the full error covariance estimated on 2018.}
\label{tab:netload}
\centering
\footnotesize
\setlength{\tabcolsep}{4pt}
\begin{tabular}{@{}lcc@{}p{1.6em}@{}lcc@{}}
\cmidrule[\heavyrulewidth]{1-3}\cmidrule[\heavyrulewidth]{5-7}
\textbf{Model} & \textbf{sMAPE (\%)} & \textbf{RMSE (MW)} & & \textbf{Model} & \textbf{sMAPE (\%)} & \textbf{RMSE (MW)} \\
\cmidrule[\lightrulewidth]{1-3}\cmidrule[\lightrulewidth]{5-7}
\multicolumn{3}{@{}>{\columncolor{black!8}[0pt][0pt]}l@{}}{\textit{Baselines}} & & \multicolumn{3}{@{}>{\columncolor{black!8}[0pt][0pt]}l@{}}{\textit{Direct GNN, one model on net-load}} \\
Persistence (1 day) & 8.89 & 5636 & & FastRGCNConv & $2.19 \pm 0.06$ & $1324 \pm 33$ \\
Chronos-Bolt & 5.70 & 3676 & & GMMConv & $2.35 \pm 0.07$ & $1401 \pm 36$ \\
ARIMA-X & 5.04 & 2933 & & RGATConv & $2.35 \pm 0.07$ & $1415 \pm 39$ \\
Chronos-2 & 3.43 & 2344 & & SuperGATConv & $2.38 \pm 0.07$ & $1421 \pm 36$ \\
XGBoost (regional) & 3.13 & 1938 & & PANConv & $2.37 \pm 0.08$ & $1435 \pm 43$ \\
GAM (national) & 2.92 & 1763 & & GCN2Conv & $2.46 \pm 0.07$ & $1460 \pm 36$ \\
XGBoost (pooled) & 2.89 & 1757 & & GATConv & $2.43 \pm 0.08$ & $1461 \pm 42$ \\
LSTM & 2.83 & 1746 & & FiLMConv & $2.52 \pm 0.07$ & $1509 \pm 41$ \\
GAM (regional) & 2.86 & 1723 & & GatedGraphConv & $2.52 \pm 0.07$ & $1524 \pm 39$ \\
XGBoost (pooled, decomposed) & 2.72 & 1678 & & ResGatedGraphConv & $2.47 \pm 0.08$ & $1541 \pm 49$ \\
TAM (regional) & 2.79 & 1659 & & & & \\
MLP (regional) & 2.48 & 1473 & & & & \\
\cmidrule[\lightrulewidth]{1-3}\cmidrule[\lightrulewidth]{5-7}
\multicolumn{3}{@{}>{\columncolor{black!8}[0pt][0pt]}l@{}}{\textit{Direct GNN, expert aggregation}} & & \multicolumn{3}{@{}>{\columncolor{black!8}[0pt][0pt]}l@{}}{\textit{Decomposed GNN, one model per component}} \\
MLpol (top-level) & $2.04 \pm 0.06$ & $1230 \pm 34$ & & FastRGCNConv (best) & $1.91 \pm 0.05$ & $1168 \pm 31$ \\
MLpol (bottom-up) & $2.03 \pm 0.06$ & $1224 \pm 33$ & & MLpol (top-level) & $1.80 \pm 0.05$ & $1109 \pm 33$ \\
Uniform average & $2.01 \pm 0.06$ & $1220 \pm 35$ & & MLpol (bottom-up) & $1.80 \pm 0.05$ & $1101 \pm 32$ \\
& & & & Uniform average & $\mathbf{1.78 \pm 0.05}$ & $\mathbf{1098 \pm 33}$ \\
\midrule
\multicolumn{7}{@{}>{\columncolor{black!8}[0pt][0pt]}l@{}}{\textit{FastRGCNConv reconciled by MinT onto MLP (regional)}} \\
Direct, diagonal & $2.11 \pm 0.06$ & $1269 \pm 33$ & & Decomposed, diagonal & $1.90 \pm 0.05$ & $1152 \pm 31$ \\
Direct, full & $2.11 \pm 0.06$ & $1269 \pm 33$ & & Decomposed, full & $1.90 \pm 0.05$ & $1151 \pm 31$ \\
\cmidrule[\heavyrulewidth]{1-3}\cmidrule[\heavyrulewidth]{5-7}
\end{tabular}
\end{table}

\begin{figure*}[!htbp]
\centering
\includegraphics[width=\textwidth]{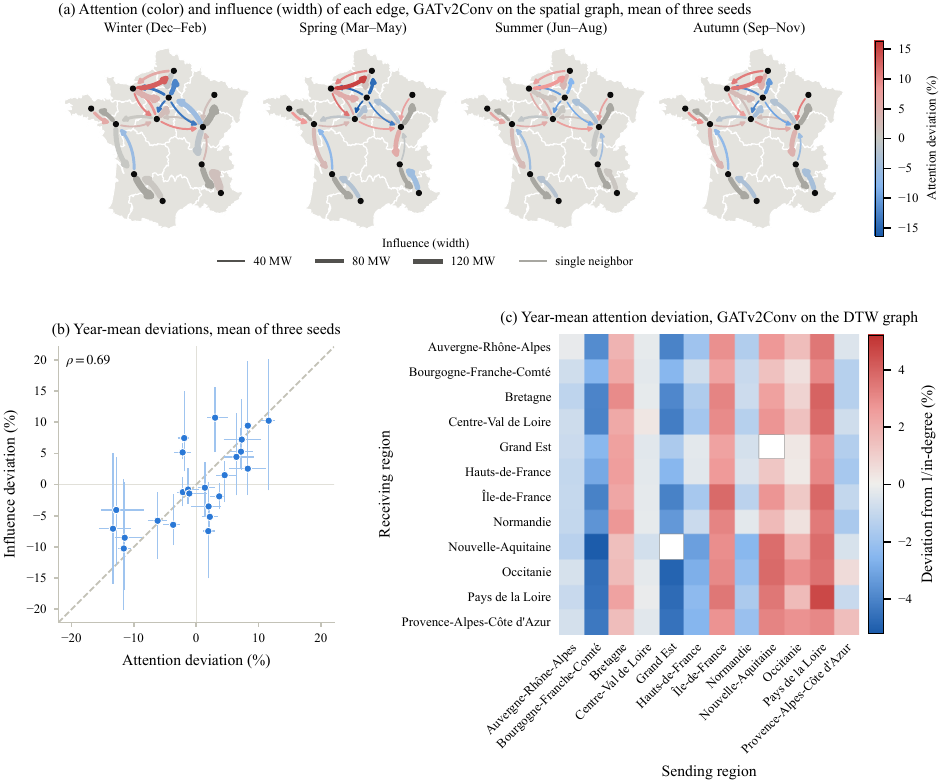}
\caption{Attention of GATs on French regional load, 2019 test set. (a) GATv2 on the spatial graph, averaged over $3$ training seeds: one arrow per directed edge, colored by the seasonal attention deviation $m_i s_{ij}-1$ of the receiving region~$i$ and with a width proportional to the occlusion influence of the sending region on the forecast of~$i$. Grey arrows enter the $4$ regions with a single neighbor, whose attention share is $1$ by construction. (b) Year-mean attention deviation against influence deviation for the $24$ edges entering regions with several neighbors, averaged over the $3$ seeds, with bars spanning the seeds, $\rho$ the Spearman correlation, and the dashed line marking equality. (c) GATv2 on the DTW graph: cell $(i,j)$ is the year-mean deviation $d_i\alpha_{ij}-1$ of the weight that region~$i$ (row) gives to region~$j$ (column). White cells are the two absent edges, and nearly identical rows show that the weight depends mostly on the sender.}
\label{fig:explain}
\end{figure*}

\subsection{Interpretability diagnostics}
\begin{figure*}[!htbp]
\centering
\includegraphics[width=\textwidth]{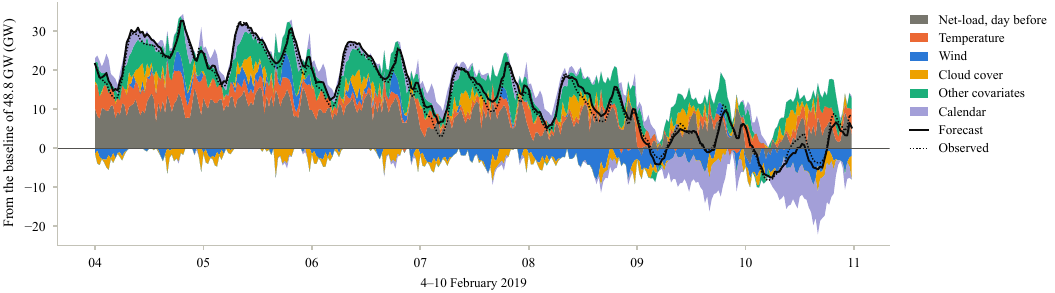}
\caption{Feature-group contributions to the national net-load forecast of the additive graph model during 4--10 February 2019, the week with the highest standard deviation of observed national wind production. Contributions are centered by their means over the 2019 test year, and the baseline contains those means and the intercept. Their sum therefore equals the forecast minus the baseline.}
\label{fig:ale}
\end{figure*}

Figure~\ref{fig:explain} plots the attention weights and occlusion influences that the interpretability module computes for GATs trained on French load ($2$ layers of width $64$, $1$ head), whose national RMSEs, from $1100$ to $1238$~MW, keep them outside the benchmark as diagnostic models. GATConv scores the edge from $j$ to $i$ by $\mathrm{LeakyReLU}(u_j+v_i)$, where $u_j$ depends on the sender and $v_i$ on the receiver. When the scores entering region~$i$ fall on the same branch of the LeakyReLU, $v_i$ cancels in the softmax and the weight $\alpha_{ij}$ depends on $i$ only through its neighborhood, the static attention identified by \citet{brody2021attentive}. The DTW graph connects every pair of regions except Grand Est and Nouvelle-Aquitaine, so the $10$ regions connected to all others have the same neighborhood and the fitted GATConv gives them identical weights, to within $4\times10^{-8}$ on every test day. Their relative deviation $d_i\alpha_{ij}-1$ from the uniform weight $1/d_i$, where $d_i$ is the in-degree of~$i$, self-loop included, then depends on the sender only. A GATv2, whose scoring function allows rankings specific to the receiver, still has $94\%$ of the variance of these deviations explained by the sender (panel~c). Passing the DTW weights to the attention as an edge attribute makes the attention weights specific to each pair, and they then follow the DTW weights (Spearman correlation $0.85$ for GATConv and $0.81$ for GATv2) without lowering the RMSE ($1157$ and $1107$~MW, against $1155$ and $1100$~MW without them).

The spatial graph links $14$ pairs of nearby regions, so the neighborhoods differ and each of its $28$ directed edges receives its own weight. An attention weight sets the share of a neighbor's message in the aggregate, whereas the effect of that neighbor on the forecast also depends on how much its message varies and on how the following layers transform the aggregate. We therefore set each weight against an occlusion influence, the mean absolute change in the forecast of region~$i$ when the inputs of region~$j$ are replaced by their test-year mean. We train the GATv2 on this graph with $3$ seeds ($42$, $0$, and $1$) and average both quantities over the $3$ trainings. A neighbor then moves the forecast by $27$ to $146$~MW and a region two edges away by about $10$~MW, against $662$~MW for the region's own inputs. Within the neighborhood of~$i$, we compare both quantities with the uniform share through $m_i s_{ij}-1$, where $m_i = d_i - 1$ is the number of neighbors of~$i$ other than itself and $s_{ij}$ the share of~$j$ in the attention or in the influence. Over the $24$ edges entering the $8$ regions with several neighbors, the year-mean deviations of attention and influence have a Spearman correlation of $0.69$ (panel~b), from $0.26$ in winter to $0.81$ in summer. The attention rankings are reproducible across trainings, with correlations of $0.93$ to $0.97$ between seeds, whereas the influence rankings vary more ($0.03$ to $0.89$), which is why we compare them on the mean over trainings. The attention of this model thus favors, on average, the neighbors with the larger effect on a region's forecast, while the effect of a given neighbor remains uncertain from one training to the next.

We examine feature attribution on net-load using the additive graph model with the same weather information as the models in Table~\ref{tab:netload}. At $1833$~MW RMSE, it is less accurate than every direct convolution in that table and is used only for interpretation. Its additive structure decomposes the national prediction into an intercept and feature-group contributions, shown in Figure~\ref{fig:ale} for 4--10 February 2019. Over the test year, the standard deviations of these contributions are $6.7$~GW for net-load lags, $4.0$~GW for calendar variables, $3.8$~GW for other covariates, $3.4$~GW for temperature, $2.9$~GW for cloud cover, and $2.6$~GW for wind. To compute the accumulated local effect of one trajectory value, we vary that value in one region while keeping all other inputs at their observed levels. The resulting standard deviation is below $14$~MW for each wind value and $7$~MW for each cloud-cover value. Each weather group thus acts through many correlated inputs, each with a small effect. The largest effects among net-load lags occur in the final $2$ hours before forecast origin, reaching $320$~MW. The most influential temperature values occur in the afternoon of the target day. This decomposition gives the additive terms of the fitted model, whereas attention weights describe only how a node aggregates its neighbors. Its attributions remain associational.

\FloatBarrier
\section{Discussion and limitations}\label{sec:discussion}
We draw four methodological conclusions from the benchmark. The evaluated graph models are shallow, with $2$ message-passing layers for load and $1$ for net-load, which keeps their computational cost modest. This design is consistent with over-smoothing analyses that predict a loss of node-specific signal as depth grows \citep{oono2020graph}, so a user should select depth on validation data like any other hyperparameter. Online expert aggregation improves on the best individual operator under the paired predictive-accuracy test, both at the national level and when applied bottom-up. Updating one graph model daily reduces RMSE by $175$~MW on average across $5$ runs, compared with a $51$~MW difference between the listed load convolutions. MinT reconciliation also improves the best load operator when combined with each tested national forecast, with larger gains for more accurate national forecasts. It improves the best direct net-load forecast when reconciled with TAM. GraphToolbox makes these alternatives available within the same experimental pipeline.

A number of limitations remain. The benchmark covers two tasks in one territory over one test year, so its findings hold in that protocol. Its fixed configuration also conditions the ranking of the operators, since the search run on $3$ of them lowers the RMSE of APPNP by $122$~MW and of GATConv by $27$~MW and raises that of LEConv by $7$~MW, and XGBoost receives a search over $10$ configurations that the graph models do not. GraphToolbox is at version 0.3.1, and its public interface may still evolve. It targets homogeneous node-level graphs and therefore excludes heterogeneous graphs and point-cloud operators. The interpretability tools inherit the known caveats of attention-based and perturbation-based attribution, and Section~\ref{sec:cases} measures this limitation through an occlusion influence that varies across training seeds, so that we compare attention and influence using the mean over several training runs. The significance analysis conditions on a single training run per architecture, so its intervals and tests quantify variability over the test window and not across retraining. The seed study of Section~\ref{sec:design} measures the second source separately, and the two would be better quantified jointly. The package ships with continuous integration and a test suite.

\section{Conclusion}\label{sec:conclusion}
GraphToolbox unifies graph construction, architectural benchmarking, online aggregation and interpretability behind one configuration-driven interface on PyG. It supports $53$ convolutional operators without modifying their implementations. In the controlled benchmark on French load, aggregation reaches $1.03\%$ MAPE, while the identity-graph ablation isolates a smaller contribution from message passing, positive for $8$ of the $10$ operators. On net-load, the graph models have a lower RMSE than the additive, recurrent and XGBoost baselines, and physical decomposition lowers it further. Pooled XGBoost given the same inputs has a lower RMSE than the best operator without a significant difference, and the significance analysis identifies no unique best architecture. A daily update lowers the $5$-run mean RMSE of GatedGraphConv from $927$ to $752$~MW, and MinT reconciliation onto the pooled XGBoost forecast lowers that of its reported run from $881$ to $747$~MW. We release the package under GPL-3.0 with documentation and examples.

\section*{Reproducibility and disclosure}
The source code, documentation, and the example notebooks underlying Section~\ref{sec:cases} are publicly available \ifblind\else from the package page on the Python Package Index, \url{https://pypi.org/project/GraphToolbox/}, and installable with \code{pip}\fi, with a lock file fixing the interpreter and every library version, the random seeds pinned, and the hardware of the benchmarks stated. A reproducibility guide maps each result to its runner. We computed the forecasts in the tables with version 0.3.0, and the interpretability diagnostics, the reconciliations and the daily updates with version 0.3.1, whose training code is unchanged. The repository carries the script that fits every baseline of Table~\ref{tab:load} and Table~\ref{tab:netload}, and the cached forecasts of both sweeps and of the identity-graph ablation, so that a user can rerun the bootstrap, the Diebold--Mariano tests and the Model Confidence Set on stored predictions, and the trained weights come as a release asset. Since the tables report one fixed set of checkpoints, retraining a row reproduces it only up to the seed spread of Table~\ref{tab:seeds}. The French load and net-load data come from RTE open data \citep{rte_ecowatt} and M\'et\'eo-France SYNOP observations \citep{meteofrance_synop}, and cover January 2015 to December 2019 at half-hourly resolution over the $12$ regions. The WEAVE-UK series of Table~\ref{tab:footprint} come from the Weave platform, in the processed release of~\citet{campagne2026weave}. In Table~\ref{tab:load}, the persistence and Chronos-Bolt rows come from a prior study on the same data \citep{campagne2025graph}, and the two additive models from \citet{campagne2026diagnostic}, the national one fitted on the aggregate load and the regional one fitted per region and summed. The TAM rows of both tables come from the TAM package \citep{tam2026package}, with the variables of the regional GAM and a temperature-weekend tensor, the regional model being selected over the national one on the 2018 validation year. We produced everything else for this paper with the scripts of the repository.\ifblind\else\space E. Campagne designed the software and the experiments and wrote the manuscript. Y. Amara-Ouali, Y. Goude, and A. Kalogeratos contributed through review and supervision.\fi\space The authors declare no competing interests.

\section*{Acknowledgment}
The authors used Claude Opus~5.5 to assist with scripts for the experiments in Section~\ref{sec:cases}, figure-generation code, and language editing. All reported numerical results were checked against the corresponding output files. The authors take responsibility for the article's content.

\bibliography{refs}
\end{document}